\documentclass[sigconf,nonacm]{acmart}

\usepackage{amsmath}
\usepackage{booktabs}
\usepackage{graphicx}
\graphicspath{{figures/}{./}}

\theoremstyle{plain}
\newtheorem{theorem}{Theorem}
\newtheorem{proposition}{Proposition}

\theoremstyle{remark}

\newcommand{\T}{\mathsf{T}}
\newcommand{\U}{\mathsf{U}}
\newcommand{\Rej}{\mathsf{R}}
\newcommand{\Loss}{\mathcal L}

\begin{document}
\emergencystretch=1.5em

\title{Does the Trust Boundary Hold over Time? A Long-Horizon Benchmark for
Receiver-Directed Federated Learning}

\author{Ting Xu}
\email{tina.xu1@ucalgary.ca}
\affiliation{%
  \institution{University of Calgary}
  \department{Department of Biomedical Engineering}
  \city{Calgary}
  \state{Alberta}
  \country{Canada}
}

\author{Henry Leung}
\email{leungh@ucalgary.ca}
\affiliation{%
  \institution{University of Calgary}
  \department{Department of Electrical and Software Engineering}
  \city{Calgary}
  \state{Alberta}
  \country{Canada}
}

\begin{abstract}
In receiver-directed federated learning, each site draws a trust boundary. For every
source and model component it judges, on its own data, whether the transfer helps
(admit), harms (reject) or cannot yet be told (abstain). We ask whether this boundary
holds over hundreds of rounds. Our benchmark has a federation of thirteen
semi-synthetic sites, an exact oracle for every evaluated transfer, paired seeds, an attack
protocol, fourteen methods over 600 rounds and Qwen3-1.7B LoRA sites over 500. The
answer depends on which boundary is measured. Across 20 paired CPU seeds, a gate
judging against a fixed reference better identified transfers useful relative to the
site's starting model. Yet it admitted 29\% of transfers harmful to the model the site
now has. A gate judging against the current model admitted 2\% of harmful transfers on
that target in the CPU tier, but missed many transfers useful to the starting model.
Two short propositions explain these distinct failures. We also evaluate a held out
audit that leaves gate verdicts intact. After training, it decides whether each final
model is deployed or the site falls back to local training, and reports admissions
that have gone stale. With independent audit data its guarantees hold after any
training history. At 600 rounds the audit deployed 65
to 72\% of models, none with negative benefit on its target, and kept 98\% of the gain.
No stale report had the wrong sign in our runs. With 20{,}000 audit observations, a
variance-adaptive bound found 99\% of stale transfers that cost at least 0.01 in log
loss. A common-target Qwen3 comparison on 60 selected transfers shows the same tension
over 100 rounds. With 38\% adversarial sites, FedFomo and CS-Gate were the only evaluated
methods with positive mean gain in every honest group.
\end{abstract}

\begin{CCSXML}
<ccs2012>
   <concept>
       <concept_id>10010147.10010257</concept_id>
       <concept_desc>Computing methodologies~Machine learning</concept_desc>
       <concept_significance>500</concept_significance>
   </concept>
   <concept>
       <concept_id>10002978.10002986</concept_id>
       <concept_desc>Security and privacy~Formal methods and theory of security</concept_desc>
       <concept_significance>300</concept_significance>
   </concept>
</ccs2012>
\end{CCSXML}
\ccsdesc[500]{Computing methodologies~Machine learning}
\ccsdesc[300]{Security and privacy~Formal methods and theory of security}

\keywords{federated learning, trust boundary, benchmark, admission control,
robustness, fairness, long-horizon evaluation, audit}

\maketitle

\section{Introduction}
Federated learning lets sites train a model together while each keeps its data. When
sites draw their data from different relationships, averaging can pull a shared
model away from what holds for a small group. Receiver-directed methods give each
site a say over what it incorporates: personalised training weights peers by the
receiver's validation loss~\cite{zhang2021fedfomo}, and three-valued admission gates
let a receiver admit, reject or abstain on each source component from its own
evidence~\cite{fedtaps}.

How should such methods be evaluated? Most federated benchmarks report accuracy after
a fixed and fairly short training run. For admission rules this is insufficient for
three reasons. First, a receiver's verdicts change the models they are evaluated on,
so the quantity being measured moves with the method. Second, benefits drift as
models train. A source that helps early can harm later. Third, the deployed model is
the product of a whole history of decisions, and its quality is not the average
quality of those decisions. A benchmark for admission therefore has to follow methods
over long horizons, report decision quality and final outcomes separately, and know
the true benefit of each transfer it evaluates. In short, it has to ask whether a receiver's
trust boundary holds over time.

Our benchmark does this on a semi-synthetic federation with exact oracle benefits for
all CPU transfers and the selected GPU transfers. Our contributions are:
\begin{itemize}
\item A fixed evaluation protocol: thirteen sites, including minority groups with little data whose
outcome relationship differs from the majority, an exact oracle, an attack rule,
paired seeds and a set of separate metrics (Section~\ref{sec:bench}).
\item Two tiers: fourteen methods with logistic and neural sites over 150 and 600
rounds on CPU, and Qwen3-1.7B LoRA sites over 50 to 500 rounds on GPU
(Section~\ref{sec:results}).
\item Three short results that make the measurements interpretable: absorption of
the benefit at the current state, overshoot under a fixed reference, and a terminal audit that
holds after any history (Section~\ref{sec:targets}).
\item An audit layer kept apart from the gate: it never changes a verdict, decides
deployment against local training, and reports stale admissions, each under a
simultaneous bound. The benchmark scores it by deployment rate, harm among deployed
models, retained gain and the precision and recall of its stale reports
(Sections~\ref{sec:targets} and~\ref{sec:results}).
\item Findings: each reference holds its own boundary and loses the other. Scored
against its own oracle the fixed reference is the more reliable in every model class.
Scored against the receiver's current model it admits many transfers that no longer
help, and the current reference ends with slightly higher final gains. Neither
reference held both boundaries in our runs, which is why the benchmark scores both
targets and audits the deployed model. With five attackers, FedFomo and CS-Gate alone
keep every honest group's mean gain above local training.
\end{itemize}

\section{Related benchmarks}\label{sec:related}
Personalised federated learning has several benchmarks. pFL-Bench~\cite{chen2022pflbench}
reports generalisation, fairness across clients and system cost. Motley~\cite{wu2022motley}
covers cross-device and cross-silo tasks with metrics for the effect of personalisation.
Matsuda et al.~\cite{matsuda2024pflbench} compare methods across vision and language tasks.
PFLlib~\cite{zhang2025pfllib} implements 37 algorithms on 24 datasets and adds a
test of gradient leakage. FLamby~\cite{terrail2022flamby} supplies naturally split
healthcare silos with a fixed round budget, and Flower Hub~\cite{gao2026flowerhub}
packages benchmarks for simulation and deployment with runtime and communication
metrics. Byzantine benchmarks, namely Blades~\cite{li2024blades},
FedSecurity~\cite{han2024fedsecurity} and ByzFL~\cite{gonzalez2025byzfl}, score
aggregators by model accuracy under attack, ByzFL by the worst case over attacks. Closest
to our question, Li et al.~\cite{li2024hetero} compare measures of whether two clients
should collaborate, judged by the test accuracy of the models they produce, and
FedMosaic~\cite{abourayya2025whom} lets each client decide whom to rely on.

These benchmarks chiefly score the trained model or the suitability of collaboration.
They do not provide an exact oracle for the benefit of each component transfer to its
receiver at each round. Table~\ref{tab:related} summarises the distinction. Our benchmark
uses such an oracle to separate verdict quality, group gains and the deployment decision.

\begin{table}[t]
\centering\small
\caption{What existing federated benchmarks score.}\label{tab:related}
\setlength{\tabcolsep}{3pt}
\begin{tabular}{p{0.27\linewidth}ccp{0.38\linewidth}}\toprule
Benchmark & Attack & \shortstack{Truth per\\transfer} & Scored\\\midrule
pFL-Bench~\cite{chen2022pflbench} & no & no & final accuracy, fairness, cost\\
Motley~\cite{wu2022motley} & no & no & final personalised accuracy\\
Matsuda et al.~\cite{matsuda2024pflbench} & no & no & final accuracy across tasks\\
PFLlib~\cite{zhang2025pfllib} & privacy & no & final accuracy, cost, leakage\\
FLamby~\cite{terrail2022flamby} & no & no & final task metric\\
Flower Hub~\cite{gao2026flowerhub} & no & no & final quality, runtime, communication\\
Li et al.~\cite{li2024hetero} & no & no & accuracy under each collaboration measure\\
Blades, FedSecurity, ByzFL & yes & no & accuracy under attack\\\midrule
This work & yes & yes & each verdict per round, group gains, audited deployment\\\bottomrule
\end{tabular}
\end{table}

\section{Setting}\label{sec:setting}
Sites $r,s\in V$, $|V|=N$, share a model split into components $j\in J$. In round $t$,
receiver $r$ trains locally to $\tilde\theta_{r,t}$ and issues verdicts
$v_{r\leftarrow s,j,t}\in\{\T,\U,\Rej\}$. With $A_{rj,t}$ the admitted sources,
\[
\theta^{\mathrm{adm}}_{r,t,j}=\tilde\theta_{r,t,j}
+\frac{\kappa}{|A_{rj,t}|}\sum_{s\in A_{rj,t}}(\tilde\theta_{s,t,j}-\tilde\theta_{r,t,j}),
\]
with $\theta^{\mathrm{adm}}_{r,t,j}=\tilde\theta_{r,t,j}$ when $A_{rj,t}$ is empty,
followed by rebasing $\theta_{r,t+1,j}=(1-\lambda)\theta^{\mathrm{adm}}_{r,t,j}+\lambda q_{j,t}$
toward the average $q_{j,t}$ of sources not withheld by $m$ corroborating rejections.
The weight $\lambda$ is leakage that bypasses the verdicts. The benchmark fixes it.

The confidence sequence gate evaluated here, CS-Gate, sets each verdict from bounded benefit observations with
an empirical Bernstein confidence sequence~\cite{waudbysmith2024,howard2021}: it admits
when the interval lies above zero, rejects when it lies below, and abstains otherwise.
Rejected triples are re-probed with probability $p$.
Each observation is one gate-validation patient's log loss difference between the
reference and the reference with the source component imported, under a label drawn
afresh from the known outcome mechanism, divided by the largest change for any patient in
the logit so that it lies in $[-1,1]$ and keeps the sign of the mean. The interval is
the predictable plug-in empirical Bernstein sequence at level $\eta/M$ with $\eta=0.05$
over all $M$ receiver--source--component triples, so its guarantee is familywise. It
covers a weighted running average of the conditional means of its observations along
the realised path, with the sequence's predictable weights~\cite{waudbysmith2024}. Each
round reads the gate-validation patients once in a fixed order and the weights change
with the evidence, so this target weights patients and rounds unequally. It is not the
patient average used by the oracle below, even when the reference and the source stop
changing, and the two can differ in sign. The benchmark therefore treats the oracle as
a separate evaluation target: it scores verdicts against it, and the certificate's
guarantee does not transfer to it. Unless stated otherwise $\kappa=\tfrac12$, $p=0.1$, $m=2$ and the gate starts
after 20 warm-up rounds of local training.

\section{What the benchmark measures}\label{sec:targets}
Let $\Loss_r$ be receiver $r$'s loss and $P_j$ the projection on component $j$. The
benefit of importing source $u$ into reference $a$ is
$\Loss_r(a)-\Loss_r(a+\kappa P_j(u-a))$. Three references give three targets: a
\emph{fixed} local model frozen after warm-up, the receiver's \emph{current} model, and
the \emph{final} deployed model. The following results explain why the benchmark
reports them separately.

\begin{proposition}[Absorption]\label{prop:absorption}
Let $\Loss_r(\theta)=\tfrac12\|\theta-w\|^2$, $a=0$, $u=w\ne0$, $0<\kappa<1$, and import
at every round, $c_{t+1}=c_t+\kappa(u-c_t)$, $c_0=0$. The benefit relative to the fixed reference is
$\Delta^A=(\kappa-\kappa^2/2)\|w\|^2$ at every round, whereas the benefit at the current
state is $\Delta^C_t=(\kappa-\kappa^2/2)(1-\kappa)^{2t}\|w\|^2\to0$.
\end{proposition}
\begin{proof}
$w-c_t=(1-\kappa)^tw$ and, for any $c$,
$\Loss_r(c)-\Loss_r(c+\kappa(w-c))=(\kappa-\kappa^2/2)\|w-c\|^2$.
\end{proof}

A gate that judges against its current model therefore loses the evidence for a
useful source because it has admitted it. A fixed reference avoids this but answers
a narrower question.

\begin{proposition}[Overshoot]\label{prop:overshoot}
In the scalar version, let $w\ne0$, $u\ne w$, $0<\kappa<1$ and import at every round,
so $c_t\to u$. The benefit relative to the fixed reference is positive if and only if
$\min(0,2w)<\kappa u<\max(0,2w)$, while the limit is worse than the local model if and only if $u$ lies
outside $[\min(0,2w),\max(0,2w)]$. Both hold when $u$ has the sign of $w$ and
$2|w|<|u|<2|w|/\kappa$.
\end{proposition}
\begin{proof}
$\Delta^A>0$ iff $|w-\kappa u|<|w|$, and $\Loss_r(u)>\Loss_r(0)$ iff $|u-w|>|w|$.
\end{proof}

With $w=1$, $u=2.5$, $\kappa=\tfrac12$ the fixed reference reports $+0.469$ at every
round, yet the second import already loses $0.352$ and the limit has loss $1.125$
against $0.5$ locally. A fixed reference certifies one step. The deployed model needs
its own audit. Let $B^T_q\in[-1,1]$ be the normalised benefit of a frozen final
candidate $q$ on audit data, with conditional mean $\mu^T_q$ given the whole history
$\mathcal H_T$.

\begin{theorem}[Terminal audit]\label{thm:terminal}
If $m$ audit observations of each of $M$ frozen candidates have conditional mean
$\mu^T_q$ given $\mathcal H_T$ and earlier audit draws, then with probability at least
$1-\eta$, simultaneously for all candidates,
$|\bar B^T_q-\mu^T_q|\le\sqrt{(2/m)\log(2M/\eta)}$, whatever the training history.
If, in addition, the audit draws of each candidate are independent and identically
distributed given $\mathcal H_T$ and the held-out audit pool, the same holds with the
empirical Bernstein radius of Maurer and Pontil~\cite{maurer2009} at level $\eta/(2M)$
per side in place of the Hoeffding radius.
Deploying the federated model only when the lower bound of its benefit over the local
model is positive, and the local model otherwise, harms no receiver with probability
at least $1-\eta$.
\end{theorem}
\begin{proof}
Conditionally on $\mathcal H_T$ the centred audit observations are martingale
differences in an interval of length two; Azuma--Hoeffding, a union bound over the $M$
candidates and integration over $\mathcal H_T$ give the bound. The empirical Bernstein
case replaces Azuma--Hoeffding by the bound of~\cite{maurer2009}, which needs independent
draws, as the audit's are given $\mathcal H_T$. On its event, every
deployed federated model has positive benefit and a fallback has zero.
\end{proof}

The audit is a held out risk control step in the spirit of Learn then
Test~\cite{angelopoulos2021ltt}, applied to the decision to deploy.

\paragraph{Gate and audit.} We treat the audit as a separate layer with a fixed contract.
The gate answers, for each transfer during training, whether it helps by the gate's own
reference, and its three-valued verdicts are never altered afterwards. The audit runs
once training stops and before anything is deployed, on audit patients the gate never
saw. It has two duties. First, it decides deployment: a receiver's federated model is
deployed only when the lower bound of its benefit over the local model is positive, and
the receiver falls back to local training otherwise. Second, it reports: every import the
gate still admits at the final round is re-tested against the receiver's final model,
and those whose upper bound is negative are reported as stale. Each duty uses the bound
of Theorem~\ref{thm:terminal} at level $\eta$ over its own family, the receivers for
deployment and all candidates (every single import, each receiver's joint import and
each final-versus-local comparison) for the report, so the two together are wrong with
probability at most $2\eta$, whatever happened during training. Splitting $\eta$ gives a
single level. The audit holds and reports. Changing the gate is left to whoever reads the
report.

\section{The benchmark}\label{sec:bench}
\paragraph{Federation.} Covariates come from two public synthetic sources, CNODES (ICD-9
diagnoses) and SyntheticMass (SNOMED codes), mapped to 13 shared variables in six
blocks (about 60{,}000 Canadian and 974 U.S. records). Seven Canadian majority sites,
three Canadian minority sites and three U.S. sites draw outcomes from a known logistic
mechanism. The minority groups differ by a sign change in one block
(\texttt{neuro\_mental}, \texttt{cardiomet}) and keep only 30 training patients each,
so majority information is useful to them. Training, gate validation, audit and test
patients are disjoint. No real outcome is used.

\paragraph{Oracle.} Because each validation patient's outcome probability $p_i$ is
known, the expected log loss difference of any transfer is computed exactly. For
receiver $r$, reference $a$ and candidate $a'=a+\kappa P_j(u-a)$ with logits $z_i$ and
$z'_i$, and expected log loss $\ell(z;p)=p\log(1+e^{-z})+(1-p)\log(1+e^{z})$,
\[
B=\frac{\frac{1}{n_r}\sum_{i}\big[\ell(z_i;p_i)-\ell(z'_i;p_i)\big]}{\max_i|z'_i-z_i|},
\]
the uniform average over validation patients of the conditional means of the gate's
observations. A triple counts as useful when
$B>10^{-3}$, harmful when $B<-10^{-3}$ and near zero otherwise. Near-zero triples are left
out of the useful and harmful denominators. Every receiver--source--component triple
($M=13\cdot12\cdot6=936$) has an oracle value in every round, recomputed from that
round's source snapshot and each reference. Group labels describe the scenario and are
never used as truth: the oracle sign of a triple can disagree with the planted sign
flip, and it changes as models train.

\paragraph{Site model.} Logistic sites take five gradient steps per round (step size
0.5, ridge $10^{-3}$). Patients are split by identifier into training (50\%), gate
validation (25\%), audit (12.5\%) and test (12.5\%) before any fitting, and scaling is
fitted on training covariates only. An MLP variant replaces each site with a
one-hidden-layer tanh network under the same splits and oracle.

\paragraph{Methods.} FedAvg~\cite{mcmahan2017}, FedProx~\cite{li2020fedprox},
Krum~\cite{blanchard2017}, trimmed mean and median~\cite{yin2018}, RFA~\cite{pillutla2022},
FLTrust~\cite{cao2021fltrust}, BOBA~\cite{bao2024boba}, Ditto~\cite{li2021ditto},
IFCA~\cite{ghosh2020ifca}, FedFomo~\cite{zhang2021fedfomo}, Fed-Equilibrium~\cite{fedequilibrium}
(cosine filtering against the coordinatewise median, then minority up-weighting at a
Pareto knee), a single-round proxy of the Fed-TAPS verdict rule~\cite{fedtaps} (a fresh
certificate from each round's evidence, without its SMT certificate), and CS-Gate with a
frozen or a current reference. Hyperparameters are chosen per seed on the honest
sites' gate-validation loss, the data the gate itself reads, and never on test data:
the proximal weight of FedProx and the regulariser of Ditto from $\{0.01,0.1,1\}$, the
number of IFCA clusters from $\{2,3\}$, and the minority weight of Fed-Equilibrium at the
knee of its validation Pareto front over $\{0.1,0.5,1,2,4,8,14\}$. FLTrust and BOBA use
100 server-held root records drawn from honest majority training data, which favours
them. Krum, trimmed mean and BOBA are told the number of attackers ($f=\max(1,a)$ for
$a$ attackers). FedFomo weights peers by each receiver's own validation loss. The
Fed-TAPS proxy recomputes a certificate from each round's evidence alone against the
current model, which is how a single-round gate behaves when it is run repeatedly.

\paragraph{Attack.} One, three or five majority sites (8, 23, 38\%) send $-2$ times their
local update, or, for personalised senders, the pre-step state plus $-2$ times the
local increment.

\paragraph{Metrics.} Reported separately, never combined: useful coverage (share of
truly beneficial triples admitted), harmful admission (share of truly harmful triples
admitted), abstention, wrong sign events (share of triples given a definite verdict
of the wrong sign, against the current-round oracle, at least once during the run; the
GPU tier also reports the share wrong at a given round), and each honest group's
final gain in test log loss over its own local model, with the worst group and the 5\%
quantile across seeds. Trajectories are recorded every 50 rounds.

\paragraph{Paired design.} A seed fixes the site data, the splits, the initial models
and the random stream, so the two references and all methods are compared on the same
federation. We report the mean over 20 seeds, the seed-level paired difference with a
bootstrap 95\% interval over seeds, and the number of seeds in which one arm is better.
Means over a grid of re-probing rates $\{0,0.1,0.5\}$ and rebasing weights
$\{0,0.1,0.3\}$ are labelled as such.

\paragraph{GPU triples.} Evaluating all 936 triples with a language model every round is
too expensive, so the GPU tier follows 60. They are chosen once per seed by a logistic
score of each receiver--source pair, copied to its four depth blocks: the 20 triples
with the most positive score, the 20 most negative and the 20 closest to zero, so each
stratum holds five pairs times four blocks and the sample covers 15 receiver--source
pairs. The logistic score only balances the selection. The oracle is
recomputed from the Qwen3 models for every triple in every round. The 60 triples are
therefore a stratified sample of clear and borderline cases, not a random sample of the
federation, and GPU rates are not comparable in level with CPU rates.

\paragraph{Tiers and reproducibility.} The CPU tier runs all fourteen methods with
logistic sites (20 seeds, 150 and 600 rounds) and the two CS-Gate references with
one-hidden-layer MLP sites. The GPU tier runs Qwen3-1.7B-Base with LoRA adapters on the
attention projections (only the $B$ matrices train, so adapter interpolation is linear),
patients rendered as text, four depth blocks as components, 60 triples stratified by
logistic benefit, five paired seeds, $\lambda=0$ and 500 rounds, plus a $\lambda=0.3$
stress test with the current reference to 250 rounds.
The GPU tier compares references, not all fourteen methods. Configurations, seeds, a run
manifest, raw per-run outputs, aggregation scripts and a CPU-only reproduction mode will
be available at \url{https://github.com/tinaxu226/trust-boundary-benchmark} once the
paper is accepted.

\section{Results}\label{sec:results}
\begin{figure}[t]
\centering
\includegraphics[width=\linewidth]{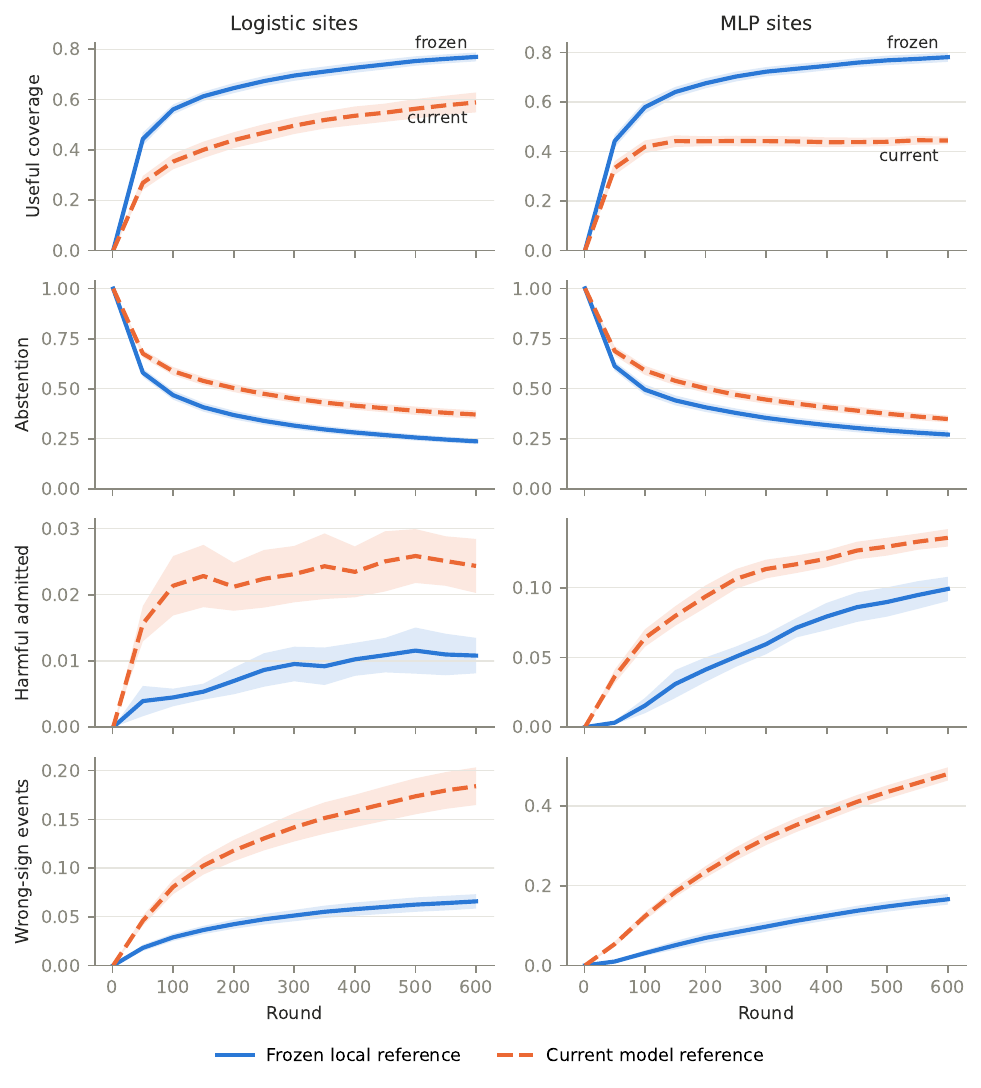}
\caption{Verdict quality over 600 rounds for logistic (left) and MLP (right) sites.
Bands are 95\% intervals across 20 seeds. Solid: frozen reference. Dashed: current.}
\label{fig:traj}
\end{figure}

\paragraph{Each reference against its own oracle.} Table~\ref{tab:ref} reports CS-Gate
with the two references, each scored against its own oracle. The frozen reference admits
more of the triples useful to it, fewer of those harmful to it and gives fewer verdicts of the
wrong sign at both horizons. Longer training mainly makes it more decisive (abstention 0.41
to 0.24). Figure~\ref{fig:traj} shows that this ordering holds at every recorded round.
With MLP sites the current reference stops gaining coverage after about 100 rounds, the
stalling that Proposition~\ref{prop:absorption} describes, and harmful admission rises
for both references as the network overfits the small minority samples and its benefits
drift.

\paragraph{Both references against a common oracle.} Scoring each arm against its own
oracle compares two different questions. We therefore scored every run of the main
setting ($p=0.1$, 600 rounds, 20 paired seeds) against both oracles on the same
trajectory. The extra scoring draws no random numbers, so the runs are unchanged.
Table~\ref{tab:cross} shows the result. For harmful admission and wrong-sign verdicts,
each reference is better on its own target in all 20 seeds. Useful coverage does not
follow that symmetry: the frozen arm is higher on the current target in 16 of 20 seeds.
Against the current-model oracle, the
question of whether an import still helps the model it is mixed into, the frozen
reference admits 29\% of harmful triples against 2\% for the current reference: the
transfers it certified against the warm-up model keep being imported after the receiver
has moved, as Proposition~\ref{prop:overshoot} predicts. The current reference holds
that boundary but covers fewer useful triples on the frozen target, as
Proposition~\ref{prop:absorption} predicts. Final test gains follow the current target:
the frozen arm's Canadian minority gain is lower by 0.030 (95\% interval 0.007 to 0.051;
lower in 15 of 20 seeds) at $\lambda=0$ and by 0.044 at $\lambda=0.3$. The GPU tier repeats
this comparison below.

\begin{table}[t]
\centering\small
\caption{CS-Gate at 600 rounds scored against both oracles on the same runs (logistic
sites, $p=0.1$, $\lambda=0$; mean over 20 paired seeds). The last column counts seeds
in which the frozen arm is better. Test gains are given in the text. $\lambda=0.3$ gives
the same pattern.}\label{tab:cross}
\setlength{\tabcolsep}{3pt}
\begin{tabular}{llccc}\toprule
Oracle & Metric & Frozen arm & Current arm & Frozen better\\\midrule
frozen & useful & 0.77 & 0.36 & 20/20\\
frozen & harmful & 0.009 & 0.052 & 20/20\\
frozen & ever wrong & 0.07 & 0.20 & 20/20\\
current & useful & 0.61 & 0.55 & 16/20\\
current & harmful & 0.287 & 0.022 & 0/20\\
current & ever wrong & 0.39 & 0.24 & 0/20\\\bottomrule
\end{tabular}
\end{table}

\begin{table}[t]
\centering\small
\caption{CS-Gate with logistic sites: mean over 20 seeds and a grid of re-probing rates
and rebasing weights.}\label{tab:ref}
\begin{tabular}{llcccc}\toprule
Rounds & Reference & Useful & Harmful & Wrong-sign & Abstain\\\midrule
150 & frozen & 0.61 & 0.005 & 0.036 & 0.41\\
150 & current & 0.40 & 0.023 & 0.102 & 0.54\\
600 & frozen & 0.77 & 0.011 & 0.066 & 0.24\\
600 & current & 0.59 & 0.024 & 0.184 & 0.37\\\bottomrule
\end{tabular}
\end{table}

\paragraph{Language model sites.} Under Qwen3 only 47--70\% of the 60 triples kept the
sign of their logistic selection score at the first gated round, so the language model
oracle, not the selection score, defines truth. Benefits kept moving for 500 rounds:
against the frozen reference 58--72\% of triples changed oracle sign at least once,
and against the current reference every triple did, because each admission changes the
model it is judged against. Over 500 rounds ($\lambda=0$, five paired seeds) the frozen
reference was better in all five paired comparisons at rounds 50, 100, 300 and 500 on
useful coverage, harmful admission and verdicts of the wrong sign. At round 500 it admitted 96\%
of the triples useful to its reference against 59\% for the current reference, and 15\%
of the triples harmful to its reference against 55\%; 8\% of decided triples had the
wrong sign at that round against 31\%, and over the whole run 27--45\% of triples were
ever given a verdict of the wrong sign against 87--97\%. The gap widened with training: under
the current reference harmful admission rose from 0.38 at round 50 to 0.55 at round 500,
while the frozen reference stayed between 0.06 and 0.24. Each arm is scored against its
own oracle here, so these numbers say how well each reference answers its own question.

To compare the references on a common target we reran the same five seeds and triples
for both references at $\lambda=0$, scored every round against both oracles, and stored
the models at rounds 25, 50, 75 and 100. The pattern of Table~\ref{tab:cross} holds: at
rounds 25, 50 and 75 each reference admitted fewer harmful triples than the other on its
own oracle in all five seeds, and at round 100 in all five on the current-model oracle
and four of five on the frozen one (in the fifth, 0.38 against 0.38). At round 100,
against the frozen oracle the frozen reference admitted 18\% of the triples harmful to it
against 43\% for the current reference. Against the current-model oracle it admitted 60\%
against 43\%. With language model sites the current reference
therefore holds its own boundary far less well than with logistic sites, and neither
reference keeps harmful admission low on the current target.

Final gains varied too much across five seeds to compare. As an additional stress test
we ran the current reference with leakage $\lambda=0.3$ to round 250. It showed no
consistent difference from $\lambda=0$ on the same seeds (harmful admission 0.53 against
0.49 at round 250). Each 500-round run took about 35 hours with up to ten runs sharing one
NVIDIA GB300, at a peak of 12\,GiB of GPU memory per run.

\begin{figure}[t]
\centering
\includegraphics[width=\linewidth]{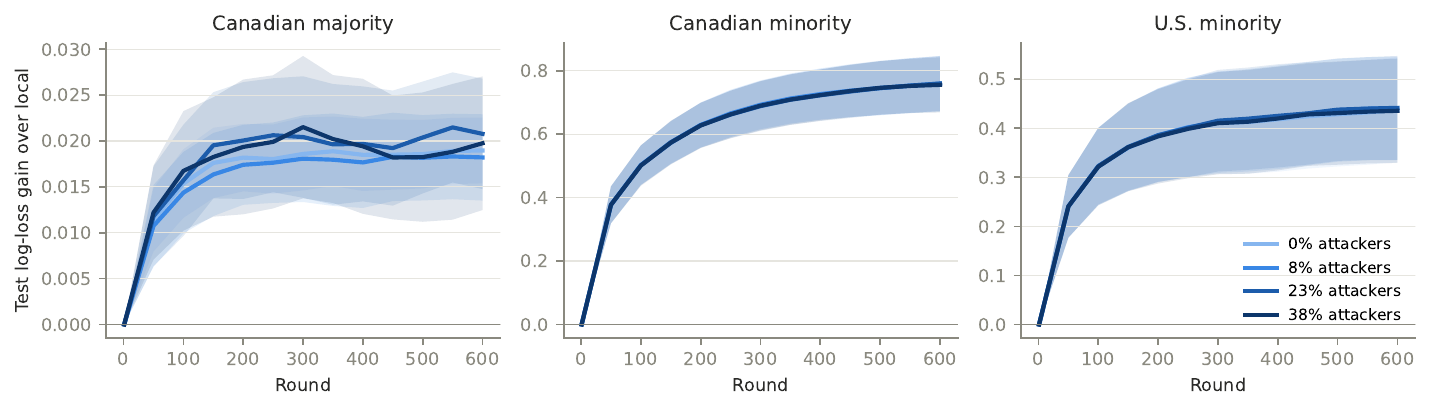}
\caption{CS-Gate's gain over the local model for each honest group at four attack
levels over 600 rounds. The curves coincide.}
\label{fig:attack}
\end{figure}

\paragraph{Baselines and attack.} Table~\ref{tab:base} gives the CPU comparison at 600
rounds. Without attackers every method except the Fed-TAPS proxy helps the minorities;
CS-Gate gives the Canadian minority the largest gain but is average for the U.S.
minority, and 14\% of its honest sites end slightly below local training, so it is not
uniformly best. With 38\% attackers, FedAvg, FedProx, Ditto, IFCA and RFA diverge, and
median, trimmed mean, Krum and Fed-Equilibrium make every group worse off. FLTrust and BOBA
remain near zero in their worst group. FedFomo and CS-Gate keep every group positive;
CS-Gate's gains are unchanged by the attack throughout training
(Figure~\ref{fig:attack}). Over 600 rounds the worst-group gain of a single seed was
slightly negative for CS-Gate in 3 of 80 seed--attack combinations (minimum $-0.006$) and
for FedFomo in 10. The Fed-TAPS proxy never admits a harmful update and never admits any
other: one round of evidence cannot certify benefits of this size, so it reduces to
local training.

\begin{table}[t]
\centering\small
\caption{Gain in test log loss over the local model for honest sites at 600 rounds,
mean over 20 seeds.}\label{tab:base}
\setlength{\tabcolsep}{3pt}
\begin{tabular}{lcccccc}\toprule
& \multicolumn{3}{c}{No attacker} & \multicolumn{3}{c}{38\% attackers}\\
\cmidrule(lr){2-4}\cmidrule(lr){5-7}
Method & CA min & US min & Worst & CA min & US min & Worst\\\midrule
FedAvg & 0.568 & 0.430 & 0.023 & $-4812$ & $-3790$ & $-6057$\\
FedProx & 0.571 & 0.440 & 0.022 & $-468$ & $-367$ & $-589$\\
Krum & 0.527 & 0.406 & 0.011 & $-1.81$ & $-1.75$ & $-2.74$\\
Trimmed mean & 0.660 & 0.478 & $-0.002$ & $-14.1$ & $-7.82$ & $-22.5$\\
Median & 0.570 & 0.449 & 0.016 & $-6.09$ & $-2.92$ & $-11.1$\\
RFA & 0.538 & 0.396 & 0.024 & $-26342$ & $-21338$ & $-33375$\\
FLTrust & 0.574 & 0.392 & 0.010 & 0.545 & 0.419 & $-0.009$\\
BOBA & 0.643 & 0.474 & 0.001 & 0.609 & 0.494 & $-0.077$\\
Ditto & 0.715 & 0.487 & 0.020 & $-3796$ & $-2931$ & $-4736$\\
IFCA & 0.662 & 0.434 & 0.022 & $-1261$ & $-713$ & $-1673$\\
FedFomo & 0.680 & 0.470 & 0.018 & 0.685 & 0.487 & 0.023\\
Fed-Equilibrium & 0.611 & 0.447 & 0.011 & $-31.2$ & $-12.0$ & $-47.5$\\
Fed-TAPS proxy & 0 & 0 & 0 & 0 & 0 & 0\\
CS-Gate & 0.758 & 0.437 & 0.019 & 0.756 & 0.436 & 0.020\\\bottomrule
\end{tabular}
\end{table}

\paragraph{Implied verdicts of weighting methods.} Methods that weight or filter sources
issue no three-valued verdicts, but each round they keep or drop every source. We read a
positive weight, membership in the kept cluster or an accepted block as admission and
anything else as rejection. These readings are a diagnostic of what each method lets in,
not certified rejections, and none of these methods can abstain. We score them against
the current-model oracle on all triples, as in Table~\ref{tab:cross}, on the 20 seeds of
the main setting (Table~\ref{tab:implied}). The classic methods are FedFomo and FLTrust.
The newer ones are pFedGraph~\cite{ye2023pfedgraph}, CoBo~\cite{hashemi2024cobo},
FreqFed~\cite{fereidooni2024freqfed} and LASA~\cite{xu2025lasa}, with
FLAME~\cite{nguyen2022flame} as a clustering reference. All are adapted versions on the
logistic sites. FLAME keeps its clustering and clipping but not its noise step. FreqFed
applies its discrete cosine transform to the 14 parameters of each update, whose order
carries no frequency meaning, so here it acts as a fixed low-dimensional projection.
LASA's layers are the six benchmark components. CoBo updates every pair in every round.
Every method uses the settings its paper gives. Where a paper gives none, as for CoBo's
penalty and weight step size, the setting is chosen per seed on the honest sites'
validation loss, as for FedProx and Ditto, over a grid of twelve. No method, including
CS-Gate, is tuned on the oracle or on test data. FedMosaic~\cite{abourayya2025whom} shares
predictions on public data rather than parameters and its weights never reach zero, so
it has no implied verdict to score.

Without attackers no other method keeps harmful admission low while admitting many
useful transfers. pFedGraph admits almost everything; FreqFed, FLAME and CoBo admit about
half of the useful and a third to a half of the harmful transfers; FedFomo keeps harmful
admission at 8\% by admitting only 18\% of the useful ones. The current-reference gate
admits 55\% and 2\%. CoBo's result is not a matter of its setting: on the same seeds,
every grid point that admitted half of the useful transfers also admitted 35--39\% of the
harmful ones, and every point that kept harmful admission below 10\% admitted at most
17\% of the useful ones. With five attackers CoBo admits 94\% of the attackers'
transfers, at least 71\% at every grid point, because scaled gradients enlarge the inner
products that set its weights, and FreqFed
admits 55\%; FedFomo, the best of the classic methods, admits 9\%. Because these methods
cannot abstain, their ever-wrong rates (48--88\% without attackers) are high by
construction and are not comparable with the gate's.

\begin{table}[tb]
\centering\small
\caption{Implied verdicts scored against the current-model oracle at round 600 (logistic
sites, 20 seeds; adapted versions, see text). Useful and harmful are the shares of truly
useful and truly harmful triples admitted. The last column is the share of the five
attackers' transfers admitted.}\label{tab:implied}
\setlength{\tabcolsep}{3pt}
\begin{tabular}{lccccc}\toprule
& \multicolumn{2}{c}{No attacker} & \multicolumn{3}{c}{38\% attackers}\\
\cmidrule(lr){2-3}\cmidrule(lr){4-6}
Method & Useful & Harmful & Useful & Harmful & Attacker\\\midrule
FedFomo (2021) & 0.18 & 0.075 & 0.14 & 0.059 & 0.09\\
FLTrust (2021) & 0.29 & 0.289 & 0.24 & 0.211 & 0.18\\
FLAME (2022) & 0.58 & 0.504 & 0.71 & 0.530 & 0.20\\\midrule
pFedGraph (2023) & 0.93 & 0.862 & 0.39 & 0.545 & 0.35\\
CoBo (2024) & 0.55 & 0.393 & 0.70 & 0.752 & 0.94\\
FreqFed (2024) & 0.69 & 0.504 & 0.63 & 0.559 & 0.55\\
LASA (2025) & 0.39 & 0.340 & 0.15 & 0.150 & 0.21\\\midrule
CS-Gate, frozen & 0.61 & 0.287 & 0.54 & 0.307 & --\\
CS-Gate, current & 0.55 & 0.022 & 0.45 & 0.033 & --\\\bottomrule
\end{tabular}
\end{table}

\paragraph{Audit.} We froze every run of the study on a common target at round 600 and
audited it with 500, 2{,}000, 5{,}000 and 20{,}000 synthetic audit observations per
candidate (Table~\ref{tab:audit}). Every simultaneous interval covered its target in
every run. With 5{,}000 observations the empirical Bernstein audit deployed 65\% of final
models in the frozen arm and 72\% in the current arm, none with negative benefit on the
audit target, and kept 98\% of the unguarded test gain. Minority sites were deployed in
93--100\% of cases and majority sites, whose gains are near 0.02, in 39--51\%. Audit size
matters: with 500 observations almost nothing was deployed, and with 20{,}000 the audit
deployed 92--95\% and kept more than 99\% of the gain. On the independent test set, which
the guarantee does not cover, the number of honest sites ending below local training fell
from 37 of 260 (frozen) and 24 (current) without the audit to 12 in each arm with 5{,}000
observations. With 20{,}000 the audit deploys more majority sites with small positive
target benefit, and 30 and 20 sites ended below local on test.

\paragraph{Stale reports.} The stale report confirms the finding on a common target from
inside the federation. At round 600 the frozen arm still admitted 7{,}958 imports over
the 20 seeds, 1{,}851 of which harm the final model, 355 of them by at least 0.01 in
log loss. In the current arm 565 of 4{,}145 were harmful, 45 by at least 0.01. No
empirical Bernstein report was wrong in sign in any run. How many stale imports the
audit finds depends on the bound. Our first version used the Hoeffding radius, which assumes every audit
patient can move the loss by the largest difference seen on the audit pool. With 5{,}000
observations it found 197 of the 1{,}851 and 140 of the 355 larger ones. When a few
patients drive the largest differences, the observed spread is far below that range.
The empirical Bernstein radius uses that spread and, with the same
5{,}000 observations, found 691 of the 1{,}851 and 316 of the 355. With 20{,}000 it found
1{,}513 and 353 of the 355. The current arm and $\lambda=0.3$ show the same pattern. What
remains unreported are mostly imports whose harm is below 0.01. A stale report is cheap
to make precise. Making it complete needs a bound that adapts to the variance and enough audit data,
and the benchmark scores both.

The GPU tier gives the same picture (Table~\ref{tab:gpuaudit}). At round 100 the audit,
with 5{,}000 observations, deployed 68\% of the 65 final models in the frozen arm and 75\%
in the current arm, none with negative benefit on the audit target, and every interval
covered its target at every stored round. More language model sites end below local
training, so falling back raised the mean test gain rather than lowering it (from 0.68 to
0.75 in the frozen arm and from 0.85 to 0.94 in the current arm) and cut the sites below
local on test from 19 to 6 and from 15 to 3. Of the 174 imports the frozen arm still
admitted, 100 harmed the final model. With 5{,}000 observations the Hoeffding radius
flagged 22 of them and the empirical Bernstein radius 65, including 64 of the 82 with harm
at least 0.01. With 20{,}000 the empirical Bernstein radius flagged 87, including 76 of the
82. In the current arm 57 of 170 were harmful. The two radii flagged 1 and 24 with 5{,}000
observations and 13 and 48 with 20{,}000. Every flag was correct in sign. The audits at
rounds 25, 50 and 75 show the same pattern: deployment between 42\% and 63\% with
5{,}000 observations, no deployed model with negative target benefit, and every flag
correct.

\begin{table}[t]
\centering\small
\caption{Audit with Qwen3 LoRA sites at round 100 ($\lambda=0$, five seeds, 13 receivers
each). Deployment uses the empirical Bernstein (EB) bound with 5{,}000 audit observations.
Stale reports are shown with the Hoeffding (H) and EB bounds, as correct reports out of
all harmful admitted imports. Gains are test log loss over local.}\label{tab:gpuaudit}
\setlength{\tabcolsep}{3pt}
\begin{tabular}{lcc}\toprule
& Frozen arm & Current arm\\\midrule
Deployed & 68\% & 75\%\\
Deployed with negative target benefit & 0 & 0\\
Test gain, no audit & 0.677 & 0.854\\
Test gain, audited & 0.746 & 0.936\\
Sites below local on test, no audit & 19/65 & 15/65\\
Sites below local on test, audited & 6/65 & 3/65\\\midrule
Admitted imports harmful at round 100 & 100/174 & 57/170\\
Stale, H, 5{,}000 / 20{,}000 obs. & 22 / 50 & 1 / 13\\
Stale, EB, 5{,}000 / 20{,}000 obs. & 65 / 87 & 24 / 48\\\bottomrule
\end{tabular}
\end{table}

\begin{table}[t]
\centering\small
\caption{Audit at round 600 (logistic sites, $\lambda=0$, 20 seeds, 13 receivers each).
Deployment uses the empirical Bernstein (EB) bound. Stale reports are shown with the
Hoeffding (H) and EB bounds, as correct reports out of all harmful admitted imports and
out of those with harm at least 0.01. Gains are test log loss over local; $\lambda=0.3$
gives the same pattern.}\label{tab:audit}
\setlength{\tabcolsep}{3pt}
\begin{tabular}{lcc}\toprule
& Frozen arm & Current arm\\\midrule
Deployed, 5{,}000 / 20{,}000 obs. & 65\% / 92\% & 72\% / 95\%\\
Deployed with negative target benefit & 0 & 0\\
Test gain, no audit & 0.286 & 0.298\\
Test gain, audited (5{,}000) & 0.281 & 0.292\\
Sites below local on test, no audit & 37/260 & 24/260\\
Sites below local on test, audited (5{,}000) & 12/260 & 12/260\\\midrule
Admitted imports harmful at round 600 & 1{,}851/7{,}958 & 565/4{,}145\\
\quad of which harm $\ge 0.01$ & 355 & 45\\
Reported, H, 5{,}000 obs. & 197; 140 & 29; 14\\
Reported, EB, 5{,}000 obs. & 691; 316 & 98; 27\\
Reported, EB, 20{,}000 obs. & 1{,}513; 353 & 381; 44\\
EB reports wrong in sign & 0 & 0\\\bottomrule
\end{tabular}
\end{table}

\section{Discussion}
\paragraph{Which boundary holds.} The answer to the title depends on which boundary is
meant. A fixed reference identifies transfers useful relative to where the receiver
started, but that target can drift away from whether an import helps the model it now
has. The current reference had fewer harmful admissions on that second target in our
paired comparisons, while missing many transfers useful on the fixed target. Neither
reference held both boundaries in our runs, and the two propositions
in Section~\ref{sec:targets} give, for each reference, a simple case in which the blind
spot comes from the reference itself rather than from the details of one gate. They are
examples, not an impossibility result, and do not rule out a rule that tracks both targets
and abstains when they disagree. The benchmark is built to score such a rule.
A benchmark that scores one target, or stops early, reports whichever reference suits
that target. This is the reason the benchmark reports both targets and treats the
deployment decision as a separate, audited step. The audit does not repair the gate.
It makes the gap visible and controls deployment risk on the audit target, using held
out observations independently of the preceding training history.

\paragraph{What long horizons reveal.} At 150 rounds, abstention is 41\% with the frozen
reference and 54\% with the current one in the grid-averaged CPU runs. By 600 rounds,
the frozen gate's abstention falls to 24\%. The current gate's useful coverage stalls
after about 100 rounds with MLP sites, while the two oracle targets diverge in the
common-target comparison and benefit signs change during the GPU runs. Short runs
would understate these changes.

\paragraph{What the benchmark does not show.} On the GPU tier the common-target
comparison and the audit cover 60 stratified triples, five seeds and 100 rounds. The 500-round GPU runs are scored on their own oracle only.
The results concern one federation and one attack, with synthetic outcomes. We do not
claim robustness to arbitrary adversaries, and the audit's guarantee covers the
synthetic audit target, not a clinical population.

\paragraph{Using the benchmark.} The CPU implementation fixes the site splits, seeds and
oracle scoring, so additional admission or aggregation rules can be evaluated under
the same protocol. The GPU tier is a long-horizon stress test for rules whose verdicts
depend on the model being trained.

\section*{Acknowledgments}
This work was supported by the Natural Sciences and Engineering Research Council of
Canada (NSERC).

\section*{GenAI Usage Disclosure}
Generative AI tools (large language models) assisted with drafting and editing prose,
proofs, experimental code and review of earlier drafts. The authors independently
reviewed all mathematical arguments, code, reported results, citations and final claims,
and take full responsibility for the manuscript and any remaining error.


\begin{thebibliography}{99}
\bibitem{fedtaps} T.~Xu, H.~Leung. Fed-TAPS: the scope of trust in federated learning
via evidence, provenance, and authority separation. Preprint, 2026.
\bibitem{fedequilibrium} T.~Xu, H.~Leung. Fed-Equilibrium framework for topological
Pareto control in robust and fair clinical federated learning. \emph{IEEE Canadian
Conference on Electrical and Computer Engineering (CCECE)}, 2026.
\bibitem{ye2023pfedgraph} R.~Ye, Z.~Ni, F.~Wu, S.~Chen, Y.~Wang. Personalized federated
learning with inferred collaboration graphs. \emph{ICML}, 2023.
\bibitem{hashemi2024cobo} D.~Hashemi, L.~He, M.~Jaggi. CoBo: collaborative learning via
bilevel optimization. \emph{NeurIPS}, 2024.
\bibitem{fereidooni2024freqfed} H.~Fereidooni, A.~Pegoraro, P.~Rieger, A.~Dmitrienko,
A.-R.~Sadeghi. FreqFed: a frequency analysis-based approach for mitigating poisoning
attacks in federated learning. \emph{NDSS}, 2024.
\bibitem{xu2025lasa} J.~Xu, Z.~Zhang, R.~Hu. Achieving Byzantine-resilient federated
learning via layer-adaptive sparsified model aggregation. \emph{WACV}, 2025.
\bibitem{nguyen2022flame} T.~D.~Nguyen, P.~Rieger, H.~Chen, H.~Yalame, H.~M\"ollering,
H.~Fereidooni, S.~Marchal, M.~Miettinen, A.~Mirhoseini, S.~Zeitouni, F.~Koushanfar,
A.-R.~Sadeghi, T.~Schneider. FLAME: taming backdoors in federated learning.
\emph{USENIX Security}, 2022.
\bibitem{zhang2021fedfomo} M.~Zhang, K.~Sapra, S.~Fidler, S.~Yeung, J.~M.~Alvarez.
Personalized federated learning with first order model optimization. \emph{ICLR}, 2021.
\bibitem{li2021ditto} T.~Li, S.~Hu, A.~Beirami, V.~Smith. Ditto: fair and robust federated
learning through personalization. \emph{ICML}, 2021.
\bibitem{ghosh2020ifca} A.~Ghosh, J.~Chung, D.~Yin, K.~Ramchandran. An efficient
framework for clustered federated learning. \emph{NeurIPS}, 2020.
\bibitem{mcmahan2017} B.~McMahan, E.~Moore, D.~Ramage, S.~Hampson, B.~A.~y~Arcas.
Communication-efficient learning of deep networks from decentralized data.
\emph{AISTATS}, 2017.
\bibitem{li2020fedprox} T.~Li, A.~K.~Sahu, M.~Zaheer, M.~Sanjabi, A.~Talwalkar, V.~Smith.
Federated optimization in heterogeneous networks. \emph{MLSys}, 2020.
\bibitem{blanchard2017} P.~Blanchard, E.~M.~El~Mhamdi, R.~Guerraoui, J.~Stainer. Machine
learning with adversaries: Byzantine tolerant gradient descent. \emph{NeurIPS}, 2017.
\bibitem{yin2018} D.~Yin, Y.~Chen, K.~Ramchandran, P.~Bartlett. Byzantine-robust
distributed learning: towards optimal statistical rates. \emph{ICML}, 2018.
\bibitem{pillutla2022} K.~Pillutla, S.~M.~Kakade, Z.~Harchaoui. Robust aggregation for
federated learning. \emph{IEEE Trans. Signal Processing}, 70:1142--1154, 2022.
\bibitem{cao2021fltrust} X.~Cao, M.~Fang, J.~Liu, N.~Z.~Gong. FLTrust: Byzantine-robust
federated learning via trust bootstrapping. \emph{NDSS}, 2021.
\bibitem{bao2024boba} W.~Bao, J.~Wu, J.~He. BOBA: Byzantine-robust federated learning with
label skewness. \emph{AISTATS}, 2024.
\bibitem{maurer2009} A.~Maurer, M.~Pontil. Empirical Bernstein bounds and sample
variance penalization. \emph{COLT}, 2009.
\bibitem{howard2021} S.~R.~Howard, A.~Ramdas, J.~McAuliffe, J.~Sekhon. Time-uniform,
nonparametric, nonasymptotic confidence sequences. \emph{Annals of Statistics},
49(2):1055--1080, 2021.
\bibitem{waudbysmith2024} I.~Waudby-Smith, A.~Ramdas. Estimating means of bounded random
variables by betting. \emph{JRSS Series B}, 86(1):1--27, 2024.
\bibitem{angelopoulos2021ltt} A.~N.~Angelopoulos, S.~Bates, E.~J.~Cand\`es, M.~I.~Jordan,
L.~Lei. Learn then test: calibrating predictive algorithms to achieve risk control.
\emph{Annals of Applied Statistics}, 19(2):1641--1662, 2025.
\bibitem{chen2022pflbench} D.~Chen, D.~Gao, W.~Kuang, Y.~Li, B.~Ding. pFL-Bench: a
comprehensive benchmark for personalized federated learning. \emph{NeurIPS Datasets and
Benchmarks Track}, 2022.
\bibitem{wu2022motley} S.~Wu, T.~Li, Z.~Charles, Y.~Xiao, Z.~Liu, Z.~Xu, V.~Smith. Motley:
benchmarking heterogeneity and personalization in federated learning. arXiv:2206.09262,
2022.
\bibitem{matsuda2024pflbench} K.~Matsuda, Y.~Sasaki, C.~Xiao, M.~Onizuka. Benchmark
for personalized federated learning. \emph{IEEE Open Journal of the Computer Society},
5:2--13, 2024. doi:10.1109/OJCS.2023.3332351.
\bibitem{zhang2025pfllib} J.~Zhang, Y.~Liu, Y.~Hua, H.~Wang, T.~Song, Z.~Xue, R.~Ma,
J.~Cao. PFLlib: a beginner-friendly and comprehensive personalized federated learning
library and benchmark. \emph{Journal of Machine Learning Research}, 2025.
\bibitem{terrail2022flamby} J.~Ogier du Terrail, S.-S.~Ayed, E.~Cyffers, et al. FLamby:
datasets and benchmarks for cross-silo federated learning in realistic healthcare
settings. \emph{NeurIPS Datasets and Benchmarks Track}, 2022.
\bibitem{gao2026flowerhub} Y.~Gao, M.~Naseri, J.~Fernandez-Marques, et al. Flower Hub: a
reproducible benchmarking platform for federated learning in simulation and deployment.
arXiv:2608.25114, 2026.
\bibitem{li2024blades} S.~Li, E.~Ngai, F.~Ye, L.~Ju, T.~Zhang, T.~Voigt. Blades: a unified
benchmark suite for Byzantine attacks and defenses in federated learning. \emph{IEEE/ACM
IoTDI}, 2024.
\bibitem{han2024fedsecurity} S.~Han, B.~Buyukates, Z.~Hu, et al. FedSecurity: a benchmark
for attacks and defenses in federated learning and federated LLMs. \emph{KDD}, 2024.
\bibitem{gonzalez2025byzfl} M.~Gonz\'alez, R.~Guerraoui, R.~Pinot, G.~Rizk, J.~Stephan,
F.~Ta\"iani. ByzFL: research framework for robust federated learning. arXiv:2505.24802,
2025.
\bibitem{li2024hetero} Z.~Li, X.~Wu, X.~Tang, T.~He, Y.-S.~Ong, M.~Chen, Q.~Liu, Q.~Lao,
H.~Yu. Benchmarking data heterogeneity evaluation approaches for personalized federated
learning. \emph{FL@FM-NeurIPS Workshop}, 2024.
\bibitem{abourayya2025whom} A.~Abourayya, J.~Kleesiek, B.~Rao, M.~Kamp. Whom to trust?
Adaptive collaboration in personalized federated learning. arXiv:2507.00259, 2025.
\end{thebibliography}
\end{document}